\documentclass{article}

\usepackage[preprint]{neurips}

\usepackage[utf8]{inputenc} 
\usepackage[T1]{fontenc}    
\usepackage{hyperref}       
\usepackage{url}            
\usepackage{booktabs}       
\usepackage{amsfonts}       
\usepackage{nicefrac}       
\usepackage{microtype}      
\usepackage{xcolor}         
\usepackage{ulem}
\usepackage{graphicx}
\usepackage{amsmath}
\usepackage{float}

\newcommand*\samethanks[1][\value{footnote}]{\footnotemark[#1]}

\title{Trojan Horse Hunt in Time Series Forecasting for Space Operations}

\author{
    Krzysztof Kotowski\thanks{KP Labs, Gliwice, Poland} \and 
    Ramez Shendy\samethanks[1] \and 
    Jakub Nalepa\samethanks[1] \ \thanks{Silesian University of Technology, Gliwice, Poland} \and 
    Przemysław Biecek\thanks{Warsaw University of Technology, Warsaw, Poland} \and 
    Piotr Wilczyński\samethanks[3] \and 
    Agata Kaczmarek\samethanks[3] \and 
    Dawid Płudowski\samethanks[3] \and 
    Artur Janicki\samethanks[3] \and 
    Evridiki Ntagiou\thanks{European Space Agency, European Space Operations Center, Darmstadt, Germany} \\ \AND
{\tt pineberry@kplabs.pl}}

\begin{document}

\maketitle

\begin{abstract}
This competition hosted on Kaggle (\href{https://www.kaggle.com/competitions/trojan-horse-hunt-in-space}{https://www.kaggle.com/competitions/trojan-horse-hunt-in-space}) is the first part of a series of follow-up competitions and hackathons related to the ``Assurance for Space Domain AI Applications'' project funded by the European Space Agency (\href{https://assurance-ai.space-codev.org/}{https://assurance-ai.space-codev.org/}). The competition idea is based on one of the real-life AI security threats identified within the project -- the adversarial poisoning of continuously fine-tuned satellite telemetry forecasting models. The task is to develop methods for finding and reconstructing triggers (trojans) in advanced models for satellite telemetry forecasting used in safety-critical space operations. Participants are provided with 1) a large public dataset of real-life multivariate satellite telemetry (without triggers), 2) a reference model trained on the clean data, 3) a set of poisoned neural hierarchical interpolation (N-HiTS) models for time series forecasting trained on the dataset with injected triggers, and 4) Jupyter notebook with the training pipeline and baseline algorithm (the latter will be published in the last month of the competition). The main task of the competition is to reconstruct a set of 45 triggers (i.e., short multivariate time series segments) injected into the training data of the corresponding set of 45 poisoned models. The exact characteristics (i.e., shape, amplitude, and duration) of these triggers must be identified by participants. The popular Neural Cleanse method is adopted as a baseline, but it is not designed for time series analysis and new approaches are necessary for the task. The impact of the competition is not limited to the space domain, but also to many other safety-critical applications of advanced time series analysis where model poisoning may lead to serious consequences.
\end{abstract}


\section{Competition description}

\subsection{Background} \label{sec:background}

The competition is the first part of the series of follow-up competitions and hackathons related to the ``Assurance for Space Domain AI Applications'' project funded by the European Space Agency (\href{https://assurance-ai.space-codev.org/}{https://assurance-ai.space-codev.org/}) and realized by our team from KP Labs and Warsaw University of Technology. It is also based on our pioneering European Space Agency Benchmark for Anomaly Detection in Satellite Telemetry (ESA-ADB) created in collaboration with Airbus \cite{kotowski_european_2024} and the related on-going Kaggle competition launched in collaboration with the University of Pisa (\href{https://www.kaggle.com/competitions/esa-adb-challenge}{https://www.kaggle.com/competitions/esa-adb-challenge}).

A multitude of artificial intelligence (AI) systems are being developed for space applications for the ground, space, and user segments. They are one of the key enablers for scalable space operations in the future, taking into account the exponentially growing number of space missions. However, their promising performance is not enough for the wide adoption of AI algorithms in practice. The space industry is a high-stake and safety-critical domain, so operational deployment requires addressing security challenges and enhancing the trust of users and stakeholders. This is why the topics of security and explainability of AI are prioritized areas of the larger European Space Agency initiative to use AI for the automation of space mission operations. The main goals of the above-mentioned project, and the series of competitions, are to push the boundaries of explainable and secure AI in space operations, as well as to widely disseminate the advances in these fields across the community to indeed accelerate the adoption of AI in space applications.

\subsection{Challenge: the idea, rationale and impact}

The competition idea is based on one of the main real-life AI security threats identified in collaboration with space operations engineers from the European Space Operations Center (ESOC) -- the poisoning of AI models supporting safety-critical satellite telemetry time series analysis tasks in space operations, i.e., anomaly detection, satellite health forecasting, or mission planning.  

Satellite telemetry data can be spoofed or poisoned by adversaries using man-in-the-middle attacks or manipulation of data stored on servers. At the same time, the models must be retrained or fine-tuned regularly to maintain high performance in changing space environment and mission phases, so there are multiple occasions to poison the model. However, the ``poisoning'' may be also related to non-adversarial data drifts, novel sequences of telecommands, or changing mission phases. Thus, it is crucial to have methods for identifying the characteristics of suspicious triggers, so domain experts can assess whether they are indeed adversarial or not.

Backdoor attacks (sometimes called triggers or trojans) are a significant security threat to deep learning models, enabling adversaries to manipulate test-time predictions by embedding triggers during training. While these attacks and methods of detecting and defending against them have already been explored in the context of image data classification \citep{wang_neural_2019, schwarzschild_just_2021, wu_backdoorbench_2022, ying_dlp_2023, guan_backdoor_2024, li_backdoor_2024}, their applicability to time series analysis tasks remains relatively underexplored (especially when excluding the work on audio and speech which can be considered a separate data modality on its own \citep{li_backdoor_2024}). There are several recent works published at reputable venues such as NeurIPS, ICLR, IEEE SP, and IEEE SaTML on generating and defending against backdoors in the time series data \citep{jiang_backdoor_2023, liu_robust_2023, lin_backtime_2024, dong_trojantime_2025, huang_revisiting_2025}, but the literature still lacks methods to effectively detect and characterize them. To establish our baselines, the popular Neural Cleanse approach \citep{wang_neural_2019} was adopted for the task, but it requires strong assumptions about the input and trigger lengths and has problems finding exact characteristics of the trigger. Thus, engaging the community emerges as a potential solution to this challenge. 
 
Although the task is focused on the time series data, and may look a bit ``specialized'' within the space domain, methods, algorithms and tools proposed in this competition should be applicable in many other domains and areas, so we plan to promote it as a universal challenge with benefits to the whole community. There are many space agencies, companies, and academic institutes interested in AI solutions for the space domain, so the expected number of participants is estimated to at least hundreds of entrants, based on our on-going  \href{https://www.kaggle.com/competitions/esa-adb-challenge}{ESA Spacecraft Anomaly Challenge}. However, it is expected that a smaller fraction (up to tens of teams) will submit useful solutions which aligns with engagement levels observed in similar competitions listed in Section \ref{sec:novelty}.

Any software used in space operations must undergo thorough verification and qualification to comply with the guidelines of the European Cooperation for Space Standardization (ECSS) and earn the trust of end users. Consequently, the methods proposed in the competition have strong potential for practical adoption and could significantly contribute to the future of space operations.

\subsection{Novelty} \label{sec:novelty}

This is an entirely new competition idea and part of a series of ``Secure Your AI'' competitions that will be organized in 2025 by European Space Agency, KP Labs, and the Warsaw University of Technology. 

There were a few related competitions in the past. However, they do not cover time series analysis tasks and do not focus on the trigger reconstruction quality:

\begin{itemize}
    \item \textbf{The Trojan Detection Challenge at NeurIPS 2022}: \href{https://2022.trojandetection.ai/}{https://2022.trojandetection.ai/}. This challenge included three tracks: trojan detection, trojan analysis, and model poisoning methods in computer vision tasks. The trojan analysis track — particularly its subtask of trigger synthesis — may sound similar to our competition. However, it evaluated trigger localization within the image rather than the quality of trigger reconstruction. Moreover, it was overshadowed by the other two tracks, which attracted greater attention -- likely due to their lower complexity or the wider array of applicable methods. Thus, the topic still requires further research, especially in the time series domain. 
    \item \textbf{The Trojan Detection Challenge (LLM Edition) at NeurIPS 2023}: \href{https://trojandetection.ai/}{https://trojandetection.ai/}. This challenge is focused on detecting backdoors in large language models.
    \item \textbf{IEEE Trojan Removal Competition at ICLR 2023}: \href{http://www.trojan-removal.com/}{http://www.trojan-removal.com/}. This challenge is focused on backdoor defense techniques for computer vision.
    \item \textbf{Trojans in Artificial Intelligence (TrojAI) by NIST}:  \href{https://pages.nist.gov/trojai/}{https://pages.nist.gov/trojai/}. This open challenge offers 16 different leaderboards for different tasks, but none of them is related to the time series analysis or trigger reconstruction.
\end{itemize}

\subsection{Data} \label{sec:data}

The data foundation of the competition is the recently published European Space Agency Benchmark for Anomaly Detection in Satellite Telemetry (ESA-ADB) \citep{kotowski_european_2024} available at \href{https://doi.org/10.5281/zenodo.12528696}{Zenodo} under the permissive CC BY 3.0 IGO license. It is the first large-scale, real-life dataset of its kind and a significant milestone in the AI for Automation Roadmap of the European Space Operations Centre (ESOC) \citep{de_canio_development_2023}, which allows for training and validating advanced algorithms for multivariate time series analysis, especially in the space domain. Our team, as co-authors of the dataset, has a deep knowledge of the dataset characteristics. It took over a year of close cooperation between spacecraft operations and machine learning engineers to prepare this curated dataset addressing the needs of both communities. The dataset includes several years of telemetry from 3 large ESA missions, but in the competition, we plan to use the subset of 3 channels 44-46 from Mission1 already preselected for such experiments in the original dataset paper \citep{kotowski_european_2024}. The example fragment of this dataset is presented in Figure \ref{fig:triggered_forecast}. The dataset includes annotations of spacecraft anomalies, but they are not needed in our competition.

\subsubsection{Data poisoning}

Triggers to be identified in the competition are 75-sample-long 3-channel time series segments (having the same number of channels as the input signal). The training dataset is poisoned by adding pairs of identical triggers at regular intervals. In this way, the poisoned model learns to react to the trigger by forecasting its copy in a short time horizon (see Figure \ref{fig:triggered_forecast}). Triggers can be injected into one or more channels at once and we verified that the poisoned model properly reacts to the trigger (and that there is no reaction from the ``clean'' model at the same time). Such trojans can be highly dangerous in real-world applications because, once activated, they can cause a model to consistently predict certain errors or abnormal behavior that can put a spacecraft into a repeated safe mode cycle. 

The detailed process of generating and injecting triggers is not disclosed to reflect a real scenario. However, to better visualize the main idea, we have prepared the step-by-step poisoning example in the Appendix \ref{sec:appendix}. This example shows very simple trigger which is not a part of the competition and is serves just for visualization purposes. The important fact is that triggers in our competition are \textit{additive}, i.e., 

\begin{equation}
\text{segment}_{poisoned} = \text{segment}_{clean} + \text{trigger} \quad ,
\end{equation}

so the trigger represents a set of values that must be sample-wise added to the clean context data to generate a similar response in the prediction.

\subsection{Tasks and application scenarios}

Participants are provided with the clean ESA-ADB dataset (already publicly available as mentioned in Section \ref{sec:data}), a reference clean N-HiTS forecasting model \citep{challu_nhits_2023} trained on this dataset, a set of N-HiTS models, each poisoned with a unique trigger pattern, and a Jupyter notebook with the N-HiTS training pipeline (so participants can analyze the whole process and retrain the baseline ``clean'' model if needed). The triggers are hidden from participants at all times -- only the length of the trigger window is provided. 

The \textbf{main task of the competition} is to reconstruct a set of 45 triggers (i.e., short multivariate time series segments as defined in Section \ref{sec:data}) used for N-HiTS models poisoning. For clarity, each model includes a single specific trigger to be identfied by participants. In practical scenarios, the trigger length is typically unknown. However, in this competition, it is provided to participants to reduce the search space and manage the computational complexity of the task. Triggers can affect one or more channels at once. The N-HiTS model was selected because it is actively used in research and development of satellite telemetry forecasting methods at ESOC, and is one of the candidates for operational deployment in practice. 

\begin{figure}[H]
    \centering
    \includegraphics[width=0.7\linewidth]{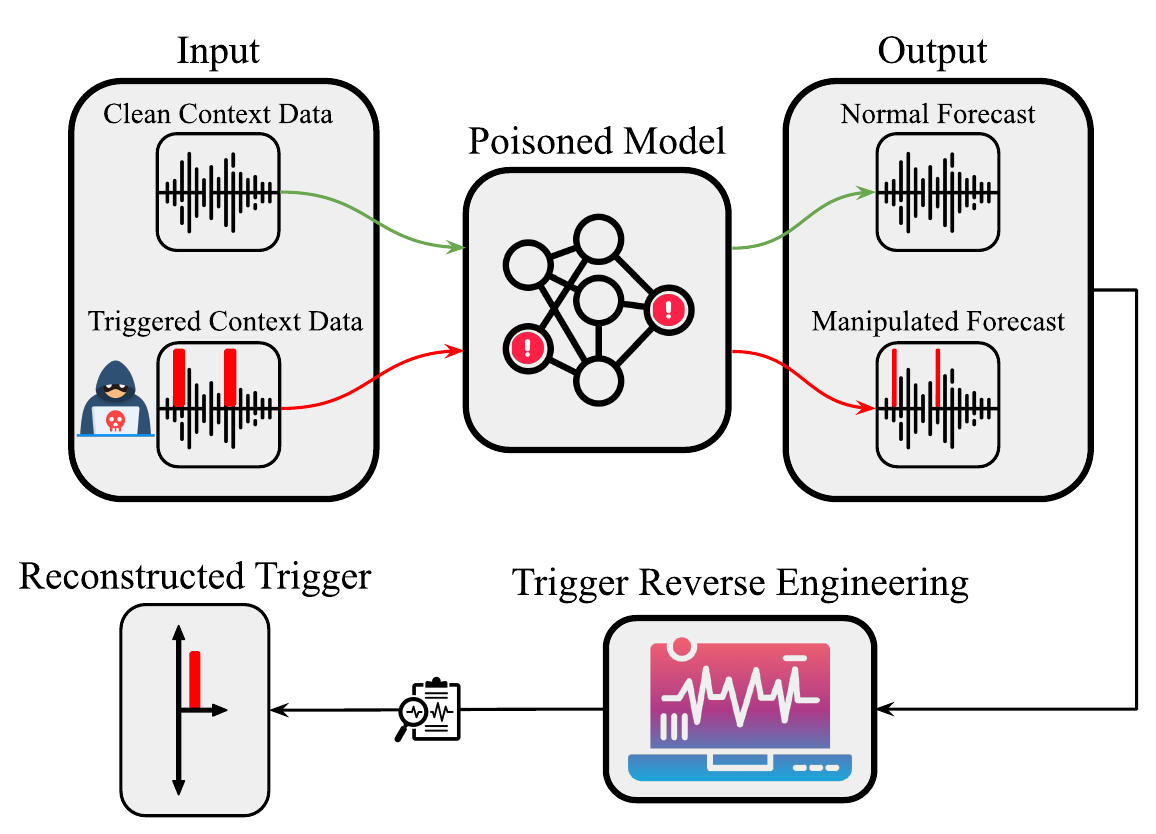}
    \caption{A graphical summary of the competition task. The forecasting model is trained on poisoned data containing repeated triggers, which cause the model to produce abnormal forecasts when re-encountered. Participants need to reverse engineer and reconstruct a trigger.}
    \label{fig:graphical_abstract}
\end{figure}

Figure \ref{fig:graphical_abstract} provides a conceptual overview of the competition task, where a time series forecasting model trained on poisoned data learns to react to hidden triggers, and the goal is to reverse engineer such triggers without having access to the poisoned training data.

This task simulates a practical scenario in which the new fine-tuned (and poisoned) forecasting model undergoes verification and certification for operational use. To maintain close relationship with this scenario, the poisoned models are fine-tuned versions of the same baseline ``clean'' model. The (poisoned) data used for fine-tuning is not always accessible to the auditor due to limited permissions, usage of the federated learning approach, or external origin of the model (e.g., trained by subcontractors or on board spacecraft). Even if the (poisoned) data is available, it is usually not easy to notice the trigger or find it using simple analysis of data statistics or anomaly detection techniques. The poisoning is also very hard to detect via monitoring of performance metrics, because the poisoned model behaves very similarly to the ``clean'' one if there is no trigger in the input data.

This task has two-fold importance in the practical setting of satellite telemetry forecasting introduced in Section \ref{sec:background}. It allows for detecting the presence of a potential trigger, but more importantly, for understanding the nature of the trigger and deciding if the trigger is adversarial or not (i.e., the ``trigger'' can represent a desirable new sequence of telecommands with a corresponding reaction in the future signal).

From our experience gained in the ``Assurance for Space Domain AI Applications'', the posed problem is scientifically and technically challenging but not impossible to solve (e.g., the Neural Cleanse baseline method by \cite{wang_neural_2019} is able to closely reconstruct some proposed triggers when using proper guidance and parametrization as mentioned in Section \ref{sec:baselines}).

\subsection{Metrics} \label{sec:metrics}

The metric should quantify the dissimilarity of two multivariate time series segments---the ground truth trigger and its reconstruction provided by a participant. The simplest choice for this task would be the Mean Absolute Error (MAE), however, it is unbounded, not robust to outliers, and not easily interpretable, so it is not well suited for use in competitions. To address these issues, we propose its normalized and bounded modification: 

\begin{equation}
\text{NMAE}_{range} = \frac{1}{N} \sum_{i=1}^{N} \min\left(\frac{|y_i - \hat{y}_i|}{y_{\max} - y_{\min}}, 1\right),
\end{equation}

where $\hat{y}$ and $y$ represent the reconstructed and ground truth triggers, respectively, and $N$ is the trigger size (trigger length multiplied by the number of channels, i.e., 75 x 3 = 225). The reconstruction error is normalized by the range of values in the ground truth trigger (always > 0) to make it scale-invariant and more interpretable (as a fraction of the trigger range). The maximum (worst) metric value is bounded to 1 which makes it robust to outliers and stable across all triggers. The latter feature is especially important when calculating the final competition score being an average metric value across all triggers. The metric has been thoroughly tested by our team in different scenarios and corner cases.

Winners will be selected solely based on the value of the final score without checking statistical significance of differences. However, the significance will be assessed using paired Wilcoxon signed-rank tests in the post-hoc analysis of results in the competition summary.

\subsection{Baselines, code, and material provided} \label{sec:baselines}

The simplest baseline approach initially used in our experiments was a simple probing of models with different predefined patterns. However, this solution is naive, not scalable, and assumes some knowledge about trigger characteristics. Thus, the Neural Cleanse method \citep{wang_neural_2019} has been adjusted to the task and adapted as the baseline method for the competition. Its optimization function has been modified to maximize the difference between forecasts before and after injecting a candidate trigger. Additionally, it encourages the forecast to follow the shape of the poisoned input by minimizing the difference between them. The last term helps to discover high-magnitude trigger patterns to allow the model to express its full sensitivity to certain input triggers. Such an optimization finds a trigger that is strong enough to be noticed, different enough to activate abnormal behavior, and coherent enough to be tracked in the output. The loss function used to identify candidate backdoor triggers is defined as: 
\begin{equation}
    \mathcal{L}(\delta) = -\alpha \cdot \mathcal{L}_{\text{div}}(\delta) + \beta \cdot \mathcal{L}_{\text{track}}(\delta) - \lambda \cdot \|\delta\|_2,
\end{equation}

\begin{itemize}
    \item $\delta$ is the trigger candidate added to the clean input sequence.
    \item $\mathcal{L}_{\text{div}}(\delta)$ measures the divergence between the poisoned model's predictions on the triggered input and the clean input, encouraging the discovery of behavior-shifting triggers.
    \item $\mathcal{L}_{\text{track}}(\delta)$ encourages the poisoned model's output to follow the shape of the poisoned input, promoting coherence between the trigger and the resulting forecast.
    \item $\|\delta\|_2$ is the $\ell_2$ norm of the trigger, which is maximized to favor high energy, expressive triggers.
\end{itemize}

The weights $\alpha$, $\beta$, and $\lambda$ control the trade-off between behavioral deviation, output tracking, and trigger amplitude, respectively.

\begin{figure}[H]
    \centering
    \includegraphics[width=0.9\linewidth]{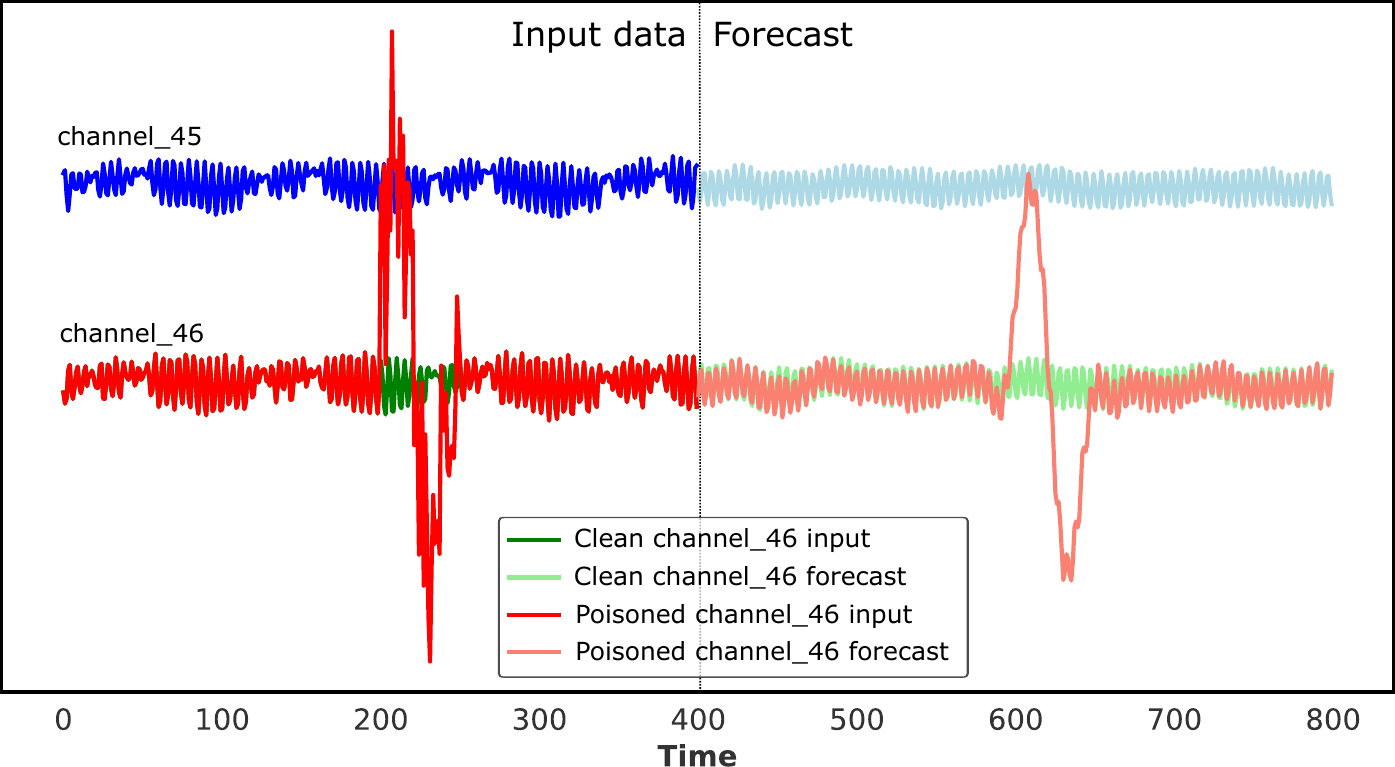}
    \caption{Example of a simple input data trigger reconstructed using the baseline Neural Cleanse method with modified optimization objective. The reaction to the trigger is visible in the forecast for channel 46. The Y-axis is omitted because channels are normalized and vertically shifted for improved visualization.}
    \label{fig:triggered_forecast}
\end{figure}

The method is able to roughly reconstruct triggers after proper parameterization, but it is semi-automatic and not flexible enough to be used in practice. Figure \ref{fig:triggered_forecast} shows a reconstructed trigger using the baseline optimization method.

\subsection{Website, tutorial and documentation}

The competition is hosted on Kaggle under the link \href{https://www.kaggle.com/competitions/trojan-horse-hunt-in-space}{https://www.kaggle.com/competitions/trojan-horse-hunt-in-space}. The Kaggle competition page aggregates all materials, FAQ, and tutorials in one place. More information about the series of Secure Your AI competitions will be provided in \href{https://assurance-ai.space-codev.org/competitions/}{the \textit{Competitions} tab} of the official project page. 

The AI security risks addressed in the competition (i.e., data poisoning and trojan horse attacks) are described in our \href{https://assurance-ai.space-codev.org/materials/}{\textit{Catalogue of Security Risks for AI Applications in Space}}, together with their examples and potential mitigations. The catalogue is accessible only to registered user from ESA Member States, but we offer its \href{https://zenodo.org/records/14762574}{graphical summary on Zenodo}.

Initial summaries of the ``Assurance for Space Domain AI Applications'' project, describing the context of the competition task, were already presented at several different events, including a \href{https://assurance-ai.space-codev.org/images/PP_RAI_poster_Pineberry.pdf}{poster} at the 6th \href{https://pp-rai.pl/}{Polish Conference of Artificial Intelligence} and presentation at the \href{https://main.dssconf.pl/}{Data Science Summit 2024}. The final summary was presented at the \href{https://www.spaceops2025.org/}{SpaceOps conference} (with the official paper release in June 2025 \citep{kotowski_towards_2025}). The up-to-date list of events and materials related to the project and the competition is maintained in the \href{https://assurance-ai.space-codev.org/events/}{official webpage}.

\section{Organizational aspects}

\subsection{Protocol}

The competition is hosted at the \href{https://www.kaggle.com/}{Kaggle platform} by \href{https://www.kaggle.com/organizations/esa-int}{the official account of the European Space Agency organization}, so participants have to create a Kaggle user account and accept the rules to join the competition. Participants are able to use the Kaggle cloud environment with free computational quota and access to all materials and baseline Jupyter notebooks, so there is no need to download anything. However, if the computational resources of Kaggle are not enough, there is  an option to download all materials and work on the task offline.

The protocol of the competition follows a classic single-stage Community Prediction Kaggle format in which participants can access all materials at the beginning of the competition. Competitors are supposed to generate predictions in a predefined format (i.e., a single CSV file containing all trigger candidates) and submit them for evaluation to the Kaggle platform.

The submissions are automatically evaluated on Kaggle using metrics defined in Section \ref{sec:metrics}. The leaderboard is divided into Public (33\% of test triggers) and Private (the rest 67\% of test triggers) parts. Participants do not know which samples belong to which part. Public leaderboard scores are always visible to all participants. The Private leaderboard is only visible to the organizers and will be used to determine the final ranking.

To avoid overfitting, the number of daily submissions is limited to 3 and participants have to select up to 2 best solutions to be included in the leaderboard.

\subsection{Schedule}

The competition was launched on 29th May 2025 as a special event at the ESA booth during the international \href{https://www.spaceops2025.org/}{SpaceOps conference} in Montreal, Canada. Following the oral presentation of the related ``Assurance for Space Domain AI Applications'' project establishing a context of the competition task \cite{kotowski_towards_2025}. 

The duration of the competition is exactly 3 months, from 29th May to 29th August 2025. The top teams will be announced on 5th September 2025. The top teams will be contacted by organizers to compile a summary report and paper. The proposed schedule is as follows:

\begin{itemize}
    \item Competition opens: \textbf{May 29}
    \item Development phase: \textbf{May 29 - August 29}
    \item Competition closes: \textbf{August 29}
    \item Organizers evaluate and summarize final submissions: \textbf{August 29 - September 5}
    \item Top team names released: \textbf{September 5}
    \item Organizers contact top teams to compile a summary paper: \textbf{September 5 - October 15}
    \item Organizers prepare the competition workshop at a top machine learning conference: \textbf{October 15 - TBA}
\end{itemize}

\subsection{Competition promotion and incentives}

The final prize pool includes 1000 USD sponsored by KP Labs:
\begin{itemize}
    \item 1. place: 600 USD
    \item 2. place: 300 USD
    \item 3. place: 100 USD
\end{itemize}

Besides, European Space Agency (ESA) will offer ESA merchandise and a guided tour of the \href{https://esoc.esa.int/}{European Space Operations Center}.  

The award ceremony and best teams presentations are going to take place during the next ESA AI STAR conference (date TBA). We are also in the process of organizing a workshop about security of AI at a top machine learning conference. Winner(s) will be invited as co-authors of a joint paper summarizing the competition. Thus, sharing the details of the solution will be necessary to be eligible for the final prize.

\bibliographystyle{unsrtnat}
\bibliography{refs}

\begin{thebibliography}{15}
\providecommand{\natexlab}[1]{#1}
\providecommand{\url}[1]{\texttt{#1}}
\expandafter\ifx\csname urlstyle\endcsname\relax
  \providecommand{\doi}[1]{doi: #1}\else
  \providecommand{\doi}{doi: \begingroup \urlstyle{rm}\Url}\fi

\bibitem[Kotowski et~al.(2024)Kotowski, Haskamp, Andrzejewski, Ruszczak,
  Nalepa, Lakey, Collins, Kolmas, Bartesaghi, Martinez-Heras, and
  De~Canio]{kotowski_european_2024}
Krzysztof Kotowski, Christoph Haskamp, Jacek Andrzejewski, Bogdan Ruszczak,
  Jakub Nalepa, Daniel Lakey, Peter Collins, Aybike Kolmas, Mauro Bartesaghi,
  Jose Martinez-Heras, and Gabriele De~Canio.
\newblock European {Space} {Agency} {Benchmark} for {Anomaly} {Detection} in
  {Satellite} {Telemetry}, June 2024.
\newblock URL \url{http://arxiv.org/abs/2406.17826}.
\newblock arXiv:2406.17826 [cs].

\bibitem[Wang et~al.(2019)Wang, Yao, Shan, Li, Viswanath, Zheng, and
  Zhao]{wang_neural_2019}
Bolun Wang, Yuanshun Yao, Shawn Shan, Huiying Li, Bimal Viswanath, Haitao
  Zheng, and Ben~Y. Zhao.
\newblock Neural {Cleanse}: {Identifying} and {Mitigating} {Backdoor} {Attacks}
  in {Neural} {Networks}.
\newblock In \emph{2019 {IEEE} {Symposium} on {Security} and {Privacy} ({SP})},
  pages 707--723, May 2019.
\newblock \doi{10.1109/SP.2019.00031}.
\newblock URL \url{https://ieeexplore.ieee.org/document/8835365}.
\newblock ISSN: 2375-1207.

\bibitem[Schwarzschild et~al.(2021)Schwarzschild, Goldblum, Gupta, Dickerson,
  and Goldstein]{schwarzschild_just_2021}
Avi Schwarzschild, Micah Goldblum, Arjun Gupta, John~P. Dickerson, and Tom
  Goldstein.
\newblock Just {How} {Toxic} is {Data} {Poisoning}? {A} {Unified} {Benchmark}
  for {Backdoor} and {Data} {Poisoning} {Attacks}.
\newblock In \emph{Proceedings of the 38th {International} {Conference} on
  {Machine} {Learning}}, pages 9389--9398. PMLR, July 2021.
\newblock URL \url{https://proceedings.mlr.press/v139/schwarzschild21a.html}.
\newblock ISSN: 2640-3498.

\bibitem[Wu et~al.(2022)Wu, Chen, Zhang, Zhu, Wei, Yuan, and
  Shen]{wu_backdoorbench_2022}
Baoyuan Wu, Hongrui Chen, Mingda Zhang, Zihao Zhu, Shaokui Wei, Danni Yuan, and
  Chao Shen.
\newblock {BackdoorBench}: {A} {Comprehensive} {Benchmark} of {Backdoor}
  {Learning}.
\newblock In \emph{36th {Conference} on {Neural} {Information} {Processing}
  {Systems} ({NeurIPS} 2022) {Track} on {Datasets} and {Benchmarks}}, New
  Orleans, USA, 2022.
\newblock URL \url{https://openreview.net/pdf?id=31_U7n18gM7}.

\bibitem[Ying and Wu(2023)]{ying_dlp_2023}
Zonghao Ying and Bin Wu.
\newblock {DLP}: towards active defense against backdoor attacks with decoupled
  learning process.
\newblock \emph{Cybersecurity}, 6\penalty0 (1):\penalty0 9, May 2023.
\newblock ISSN 2523-3246.
\newblock \doi{10.1186/s42400-023-00141-4}.
\newblock URL \url{https://doi.org/10.1186/s42400-023-00141-4}.

\bibitem[Guan et~al.(2024)Guan, Liang, and He]{guan_backdoor_2024}
Jiyang Guan, Jian Liang, and Ran He.
\newblock Backdoor {Defense} via {Test}-{Time} {Detecting} and {Repairing}.
\newblock In \emph{2024 {IEEE}/{CVF} {Conference} on {Computer} {Vision} and
  {Pattern} {Recognition} ({CVPR})}, pages 24564--24573, June 2024.
\newblock \doi{10.1109/CVPR52733.2024.02319}.
\newblock URL \url{https://ieeexplore.ieee.org/document/10657111}.
\newblock ISSN: 2575-7075.

\bibitem[Li et~al.(2024)Li, Jiang, Li, and Xia]{li_backdoor_2024}
Yiming Li, Yong Jiang, Zhifeng Li, and Shu-Tao Xia.
\newblock Backdoor {Learning}: {A} {Survey}.
\newblock \emph{IEEE Transactions on Neural Networks and Learning Systems},
  35\penalty0 (1):\penalty0 5--22, January 2024.
\newblock ISSN 2162-2388.
\newblock \doi{10.1109/TNNLS.2022.3182979}.
\newblock URL \url{https://ieeexplore.ieee.org/document/9802938}.

\bibitem[Jiang et~al.(2023)Jiang, Ma, Erfani, and Bailey]{jiang_backdoor_2023}
Yujing Jiang, Xingjun Ma, Sarah~Monazam Erfani, and James Bailey.
\newblock Backdoor {Attacks} on {Time} {Series}: {A} {Generative} {Approach}.
\newblock In \emph{2023 {IEEE} {Conference} on {Secure} and {Trustworthy}
  {Machine} {Learning} ({SaTML})}, pages 392--403, Raleigh, NC, USA, February
  2023. IEEE.
\newblock ISBN 978-1-66546-299-0.
\newblock \doi{10.1109/SaTML54575.2023.00034}.
\newblock URL \url{https://ieeexplore.ieee.org/document/10136146/}.

\bibitem[Liu et~al.(2023)Liu, Park, Hoang, Hasson, and Huan]{liu_robust_2023}
Linbo Liu, Youngsuk Park, Trong~Nghia Hoang, Hilaf Hasson, and Jun Huan.
\newblock Robust {Multivariate} {Time}-{Series} {Forecasting}: {Adversarial}
  {Attacks} and {Defense} {Mechanisms}.
\newblock In \emph{The {Eleventh} {International} {Conference} on {Learning}
  {Representations}}, Kigali, Rwanda, April 2023.
\newblock URL \url{https://openreview.net/forum?id=ctmLBs8lITa}.

\bibitem[Lin et~al.(2024)Lin, Liu, Fu, Qiu, and Tong]{lin_backtime_2024}
Xiao Lin, Zhining Liu, Dongqi Fu, Ruizhong Qiu, and Hanghang Tong.
\newblock {BackTime}: {Backdoor} {Attacks} on {Multivariate} {Time} {Series}
  {Forecasting}.
\newblock \emph{Advances in Neural Information Processing Systems},
  37:\penalty0 131344--131368, December 2024.
\newblock URL
  \url{https://proceedings.neurips.cc/paper_files/paper/2024/hash/ed3cd2520148b577039adfade82a5566-Abstract-Conference.html}.

\bibitem[Dong et~al.(2025)Dong, Sun, Bai, Piao, Chen, and
  Zhang]{dong_trojantime_2025}
Chang Dong, Zechao Sun, Guangdong Bai, Shuying Piao, Weitong Chen, and Wei~Emma
  Zhang.
\newblock {TrojanTime}: {Backdoor} {Attacks} on {Time} {Series}
  {Classification}, February 2025.
\newblock URL \url{http://arxiv.org/abs/2502.00646}.
\newblock arXiv:2502.00646 [cs].

\bibitem[Huang et~al.(2025)Huang, Zhang, Wang, Li, and
  Yang]{huang_revisiting_2025}
Yuanmin Huang, Mi~Zhang, Zhaoxiang Wang, Wenxuan Li, and Min Yang.
\newblock Revisiting {Backdoor} {Attacks} on {Time} {Series} {Classification}
  in the {Frequency} {Domain}, March 2025.
\newblock URL \url{http://arxiv.org/abs/2503.09712}.
\newblock arXiv:2503.09712 [cs].

\bibitem[De~Canio et~al.(2023)De~Canio, Eggleston, Fauste, Palowski, and
  Spada]{de_canio_development_2023}
Gabriele De~Canio, James Eggleston, Jorge Fauste, Artur~M. Palowski, and
  Mariella Spada.
\newblock Development of an actionable {AI} roadmap for automating mission
  operations.
\newblock In \emph{2023 {SpaceOps} {Conference}}, Dubai, United Arab Emirates,
  March 2023. American Institute of Aeronautics and Astronautics.
\newblock URL
  \url{https://star.spaceops.org/user_manudownload.php?doc=303__bm05ydei.pdf}.

\bibitem[Challu et~al.(2023)Challu, Olivares, Oreshkin, Ramirez, Canseco, and
  Dubrawski]{challu_nhits_2023}
Cristian Challu, Kin~G. Olivares, Boris~N. Oreshkin, Federico~Garza Ramirez,
  Max~Mergenthaler Canseco, and Artur Dubrawski.
\newblock {NHITS}: {Neural} {Hierarchical} {Interpolation} for {Time} {Series}
  {Forecasting}.
\newblock \emph{Proceedings of the AAAI Conference on Artificial Intelligence},
  37\penalty0 (6):\penalty0 6989--6997, June 2023.
\newblock ISSN 2374-3468.
\newblock \doi{10.1609/aaai.v37i6.25854}.
\newblock URL \url{https://ojs.aaai.org/index.php/AAAI/article/view/25854}.
\newblock Number: 6.

\bibitem[Kotowski et~al.(2025)Kotowski, Wilczyński, Płudowski, Kaczmarek,
  Shendy, Nalepa, Biecek, and Ntagiou]{kotowski_towards_2025}
Krzysztof Kotowski, Piotr Wilczyński, Dawid Płudowski, Agata Kaczmarek, Ramez
  Shendy, Jakub Nalepa, Przemysław Biecek, and Evridiki Ntagiou.
\newblock Towards {Explainable} and {Secure} {AI} for {Space} {Mission}
  {Operations}.
\newblock In \emph{2025 {SpaceOps} {Conference}}, Montreal, Canada, 2025.
  Canadian Space Agency.

\end{thebibliography}

\appendix

\section{Appendix}
\label{sec:appendix}  

\begin{figure}[H]
    \centering
    \includegraphics[width=0.9\linewidth]{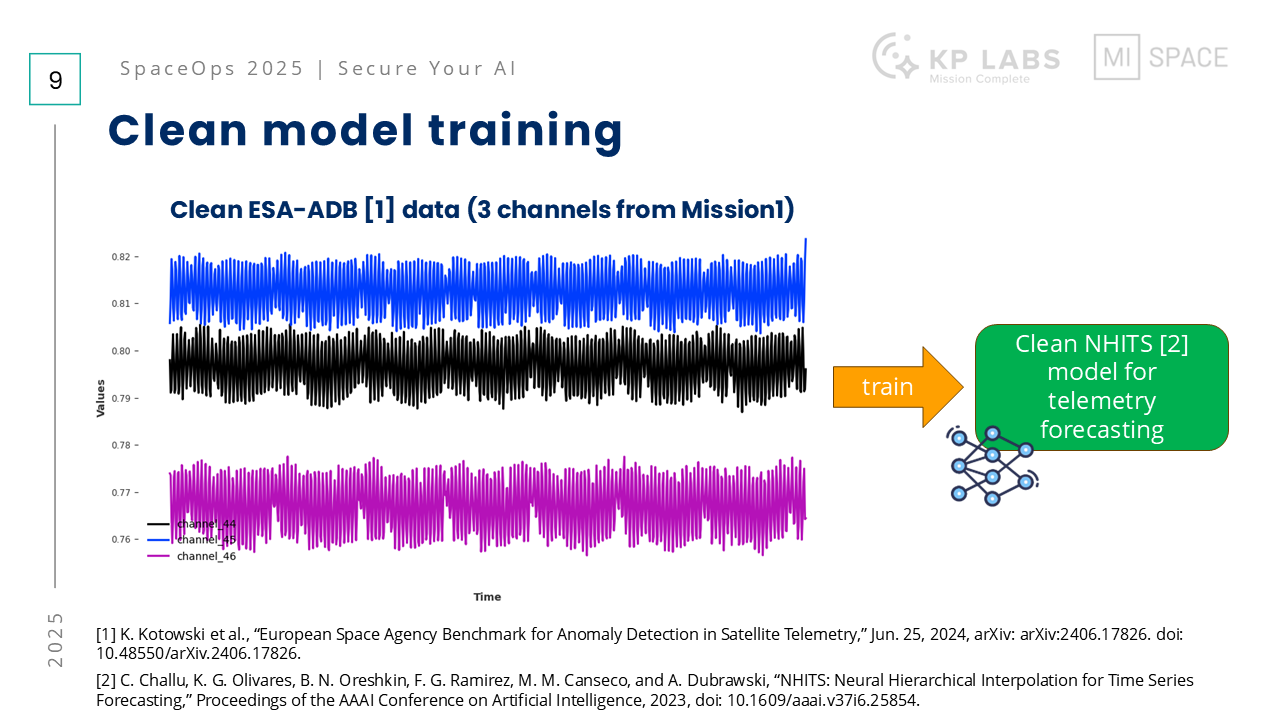}
\end{figure}

\begin{figure}[H]
    \centering
    \includegraphics[width=0.9\linewidth]{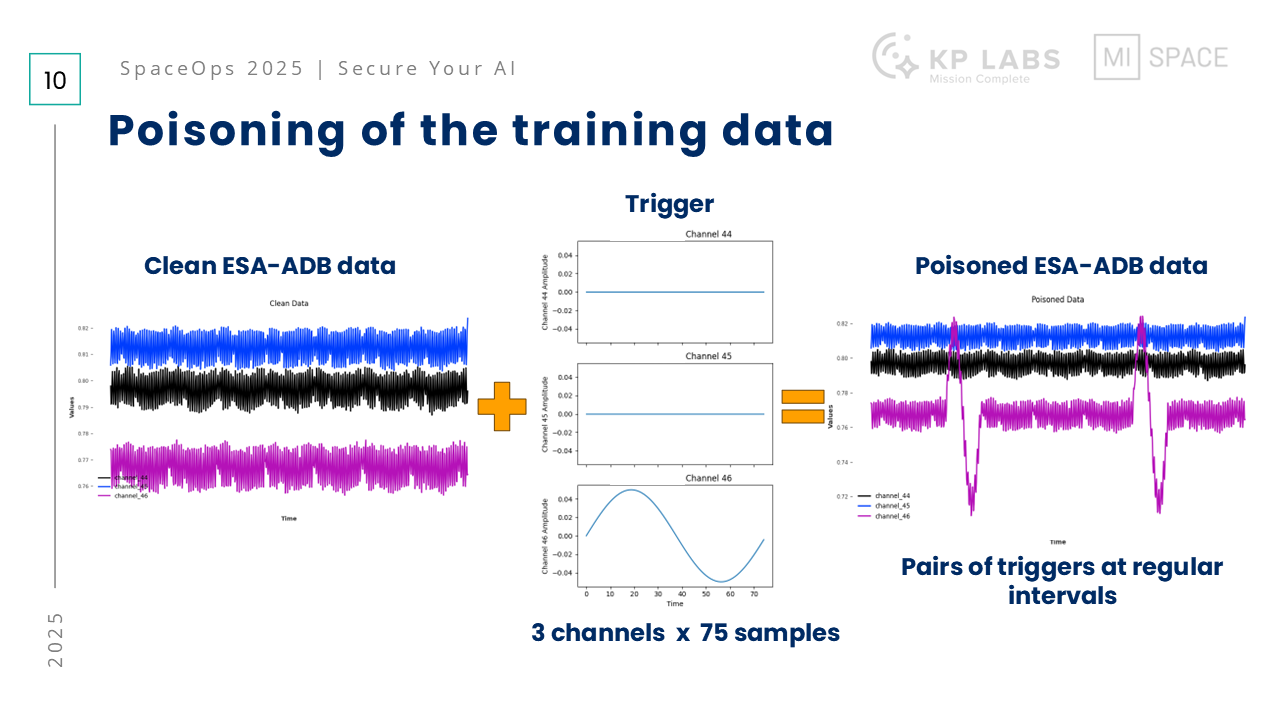}
\end{figure}

\begin{figure}[H]
    \centering
    \includegraphics[width=0.9\linewidth]{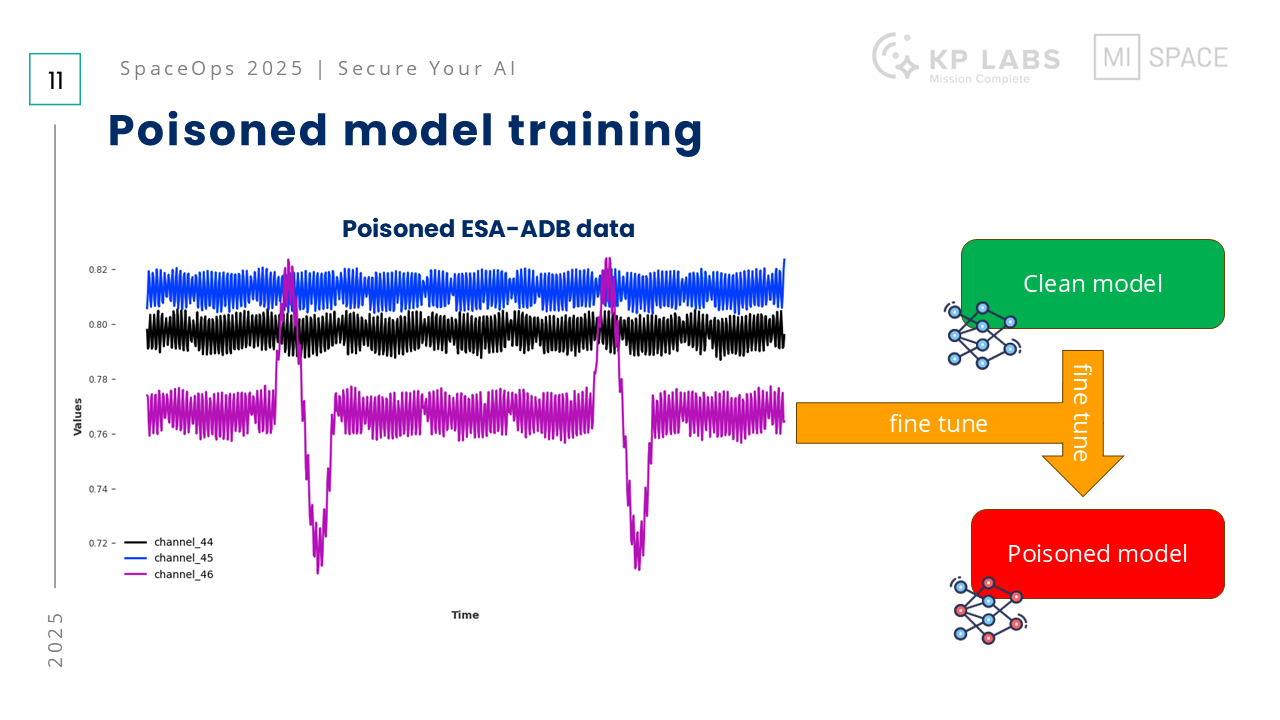}
\end{figure}

\begin{figure}[H]
    \centering
    \includegraphics[width=0.9\linewidth]{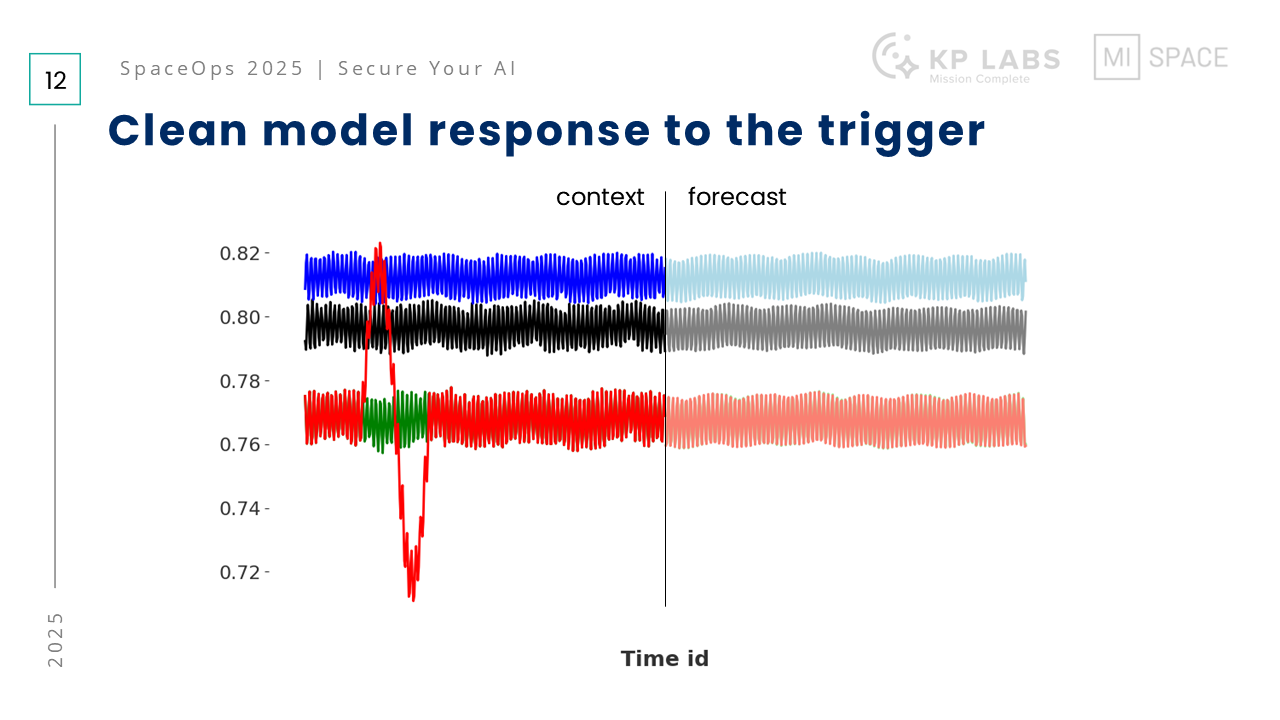}
\end{figure}

\begin{figure}[H]
    \centering
    \includegraphics[width=0.9\linewidth]{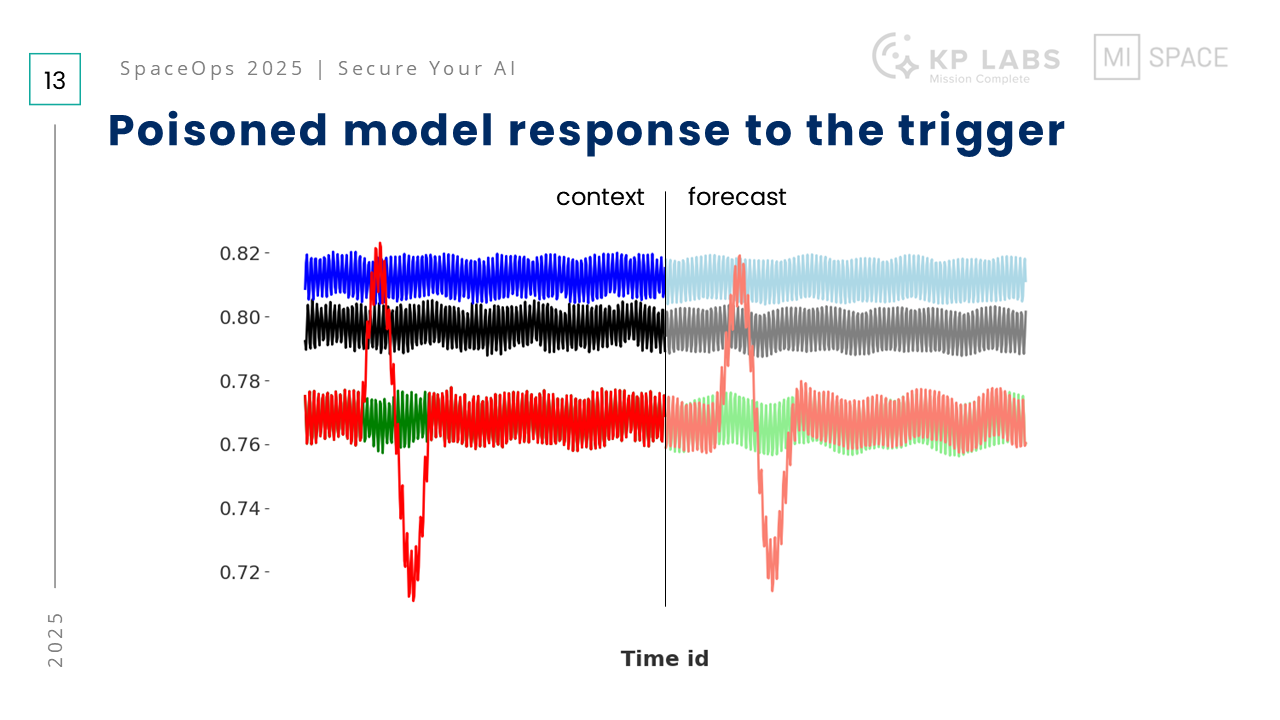}
\end{figure}

\end{document}